%% file: template.tex
\documentclass{article}

\usepackage{arxiv}

\usepackage[utf8]{inputenc} 
\usepackage[T1]{fontenc}    
\usepackage{hyperref}       
\usepackage{url}            
\usepackage{booktabs}       
\usepackage{amsfonts}       
\usepackage{nicefrac}       
\usepackage{microtype}      
\usepackage{lipsum}         
\usepackage{graphicx}
\usepackage{natbib}
\usepackage{doi}
\usepackage{amsmath}
\usepackage{amssymb}
\usepackage{booktabs}
\usepackage{multirow}
\usepackage{graphicx}
\usepackage{cleveref}       
\usepackage{float}
\usepackage{algorithm}
\usepackage{algorithmic}

\title{CS-GBA: A Critical Sample-based Gradient-guided Backdoor Attack for Offline Reinforcement Learning}

\newif\ifuniqueAffiliation
\uniqueAffiliationtrue

\ifuniqueAffiliation 
\author{
    Yuanjie Zhao\thanks{Equal contribution} \\
    SJTU Paris Elite Institute of Technology \\
    Shanghai Jiao Tong University \\
    Shanghai, China \\
    \texttt{silencezyj@sjtu.edu.cn} \\
    \And
    Junnan Qiu\thanks{Equal contribution} \\
    SJTU Paris Elite Institute of Technology \\
    Shanghai Jiao Tong University \\
    Shanghai, China \\
    \texttt{qiujunnan@sjtu.edu.cn} \\
    \And
    Yue Ding\thanks{Correspondence to: dingyue@sjtu.edu.cn} \\
    Department of Computer Science \\
    Shanghai Jiao Tong University \\
    Shanghai, China \\
    \texttt{dingyue@sjtu.edu.cn} \\
    \And
    Jie Li\thanks{Correspondence to: lijiecs@sjtu.edu.cn} \\
    Department of Computer Science \\
    Shanghai Jiao Tong University \\
    Shanghai, China \\
    \texttt{lijiecs@sjtu.edu.cn} \\
}
\else
\usepackage{authblk}

\author[1]{Yuanjie Zhao}
\author[1]{Junnan Qiu}
\author[2]{Yue Ding}
\author[2]{Jie Li}
\affil[1]{SJTU Paris Elite Institute of Technology, Shanghai Jiao Tong University}
\affil[2]{Department of Computer Science and Engineering, Shanghai Jiao Tong University}
\fi

\renewcommand{\headeright}{Technical Report}
\renewcommand{\undertitle}{Technical Report}
\renewcommand{\shorttitle}{Critical Sample-based Gradient-guided Backdoor Attack for Offline RL}

\hypersetup{
pdftitle={A template for the arxiv style},
pdfsubject={q-bio.NC, q-bio.QM},
pdfauthor={David S.~Hippocampus, Elias D.~Striatum},
pdfkeywords={Offline Reinforcement Learning, Data Poisoning, Adversarial Attacks, Global Budget Allocation, Temporal Difference Error},
}

\begin{document}
\maketitle

\begin{abstract}
    Offline Reinforcement Learning (RL) enables policy optimization from static datasets but is inherently vulnerable to backdoor attacks. Existing attack strategies typically struggle against safety-constrained algorithms (e.g., CQL) due to inefficient random poisoning and the use of easily detectable Out-of-Distribution (OOD) triggers. In this paper, we propose \textbf{CS-GBA} (\textbf{C}ritical \textbf{S}ample-based \textbf{G}radient-guided \textbf{B}ackdoor \textbf{A}ttack), a novel framework designed to achieve high stealthiness and destructiveness under a strict budget. Leveraging the theoretical insight that samples with high Temporal Difference (TD) errors are pivotal for value function convergence, we introduce an adaptive \textbf{Critical Sample Selection} strategy that concentrates the attack budget on the most influential transitions. To evade OOD detection, we propose a \textbf{Correlation-Breaking Trigger} mechanism that exploits the physical mutual exclusivity of state features (e.g., 95th percentile boundaries) to remain statistically concealed. Furthermore, we replace the conventional label inversion with a \textbf{Gradient-Guided Action Generation} mechanism, which searches for worst-case actions within the data manifold using the victim Q-network's gradient. Empirical results on D4RL benchmarks demonstrate that our method significantly outperforms state-of-the-art baselines, achieving high attack success rates against representative safety-constrained algorithms with a minimal \textbf{5\%} poisoning budget, while maintaining the agent's performance in clean environments.
\end{abstract}


\input{contents/motivation}
\input{contents/related_work}
\input{contents/theory}
\input{contents/experiment}
\input{contents/conclusion}

\bibliographystyle{unsrtnat}
\bibliography{references}



\end{document}

%% file: contents/motivation.tex
\section{Introduction}

Deep Reinforcement Learning (DRL) has achieved remarkable success in solving complex decision-making tasks, ranging from video games \cite{mnih2013playing, silver2016mastering} to robotic control \cite{lillicrap2015continuous, levine2016end}. However, the deployment of DRL in real-world scenarios is often hindered by the prohibitively high cost and safety risks associated with online interaction. To address this, Offline Reinforcement Learning (Offline RL) has emerged as a promising paradigm, enabling agents to learn optimal policies entirely from static, pre-collected datasets without interacting with the environment \cite{levine2020offline, fujimoto2021minimalist}. This data-driven approach has facilitated applications in safety-critical domains such as autonomous driving \cite{kiran2021deep} and healthcare \cite{gottesman2019guidelines}, supported by the recent emergence of large-scale robotic datasets \cite{o2024open, walke2023bridgedata}.

Despite its potential, the reliance on static datasets exposes Offline RL to severe security threats, particularly \textit{backdoor attacks} (or trojan attacks) \cite{gu2019badnets, liu2018trojaning}. In this setting, an adversary injects a small fraction of poisoned data into the training set. The victim agent behaves normally in clean environments but executes malicious actions immediately upon detecting a specific trigger pattern \cite{kiourti2020trojdrl, wang2021backdoorl}. Given that offline datasets are often aggregated from crowdsourced or third-party sources \cite{fu2020d4rl, gulcehre2020rl}, ensuring data integrity is extremely challenging, making such attacks a practical and imminent threat. Recent studies have demonstrated the feasibility of these attacks in offline settings, proposing methods to embed triggers into trajectories \cite{gong2024baffle, cui2024badrl} or tamper with rewards \cite{xu2024universal, rathbun2024adversarial}.

However, crafting effective backdoor attacks against Offline RL is significantly more difficult than in standard supervised learning or online RL. The primary challenge lies in the \textit{conservatism} mechanism inherent in state-of-the-art offline algorithms. Methods like CQL \cite{kumar2020conservative}, IQL \cite{kostrikov2021offline}, and TD3+BC \cite{fujimoto2021minimalist} explicitly penalize or constrain actions that deviate from the behavioral distribution to mitigate extrapolation errors. Consequently, existing attack methods face two major limitations: (1) \textbf{Inefficiency:} Traditional random poisoning strategies or simple trajectory modifications \cite{ma2019policy} waste the attack budget on samples that have little impact on the value function learning; (2) \textbf{Detectability:} Triggers that rely on extreme values or Out-of-Distribution (OOD) patterns are easily identified and suppressed by the conservatism regularizers (e.g., CQL's Q-value penalty), rendering the attack ineffective, as noted in recent in-distribution studies \cite{ashcraft2025backdoors}.

To bridge this gap, we propose \textbf{CS-GBA} (Critical Sample-based Gradient-guided Backdoor Attack), a novel framework designed to bypass conservative defenses while maintaining high attack efficiency. Our key insight is that not all transitions are equally important for policy learning; samples with high Temporal Difference (TD) errors contribute more significantly to the value function update \cite{schaul2015prioritized}. Based on this, we design a critical sample selection mechanism to pinpoint the most vulnerable data points. Furthermore, instead of using conspicuous triggers, we introduce a correlation-breaking trigger that exploits the physical mutual exclusivity of features to evade OOD detection. Finally, rather than simple label flipping, we utilize a gradient-guided mechanism to generate adversarial actions that minimize the agent's performance while remaining within the data manifold.

Our main contributions are summarized as follows:
\begin{itemize}
    \item We propose CS-GBA, a stealthy and efficient backdoor attack framework tailored for Offline RL, which effectively bypasses the OOD detection mechanisms of conservative algorithms.
    \item We introduce a TD-error-based sample selection strategy that significantly improves attack efficiency, allowing for successful attacks with a poisoning budget as low as 5\%.
    \item We design a gradient-guided action generation method and a correlation-breaking trigger, ensuring both the destructiveness of the attack and the preservation of clean performance.
    \item Extensive experiments on D4RL benchmarks \cite{fu2020d4rl} demonstrate that CS-GBA outperforms state-of-the-art baselines \cite{gong2024baffle, cui2024badrl} against representative algorithms including CQL \cite{kumar2020conservative}, IQL \cite{kostrikov2021offline}, and BCQ \cite{fujimoto2019off}.
\end{itemize}

%% file: contents/related_work.tex
\section{Related Work}

\textbf{Offline Reinforcement Learning.}
Offline RL aims to learn optimal policies entirely from static datasets without further interaction with the environment \cite{levine2020offline}. To mitigate the extrapolation error caused by distribution shifts, representative algorithms like CQL \cite{kumar2020conservative}, IQL \cite{kostrikov2021offline}, and TD3+BC \cite{fujimoto2021minimalist} incorporate conservatism mechanisms. Specifically, CQL minimizes the Q-values of Out-of-Distribution (OOD) actions, BCQ \cite{fujimoto2019off} constrains the policy to select actions similar to those in the behavior policy, and IQL avoids querying unseen actions by treating value estimation as an expectile regression problem. Although these defenses are effective against natural distribution shifts, they implicitly assume the dataset is benign. This reliance on data integrity makes them inherently vulnerable to malicious data poisoning, where a small fraction of adversarial data can manipulate the learned policy.

\textbf{Adversarial Attacks on RL.}
The vulnerability of RL agents has been extensively studied in various settings. In online scenarios, adversaries can perturb observations during the testing phase to mislead the agent \cite{gleave2019adversarial, huang2017adversarial}.
In the offline setting, the threat is magnified as the agent cannot correct malicious biases through interaction. Early works in backdoor attacks demonstrated that injecting trigger patterns into training data can control model behavior \cite{gu2019badnets, kiourti2020trojdrl}.
Recently, research has shifted towards more sophisticated offline-specific attacks. For instance, \textbf{BadRL} \cite{cui2024badrl} targets specific states to implant backdoors, while other works explore reward tampering \cite{xu2024universal, rathbun2024adversarial} or temporal-pattern triggers \cite{yu2022temporal}. However, applying these attacks directly to robust Offline RL algorithms remains challenging, as the conservative nature of these algorithms (e.g., OOD detection) can effectively filter out naive poisoning attempts.

\textbf{Stealthiness and Efficiency Challenges.}
To evade detection, recent research focuses on \textit{stealthy} attacks that maintain the agent's performance in clean environments. Methods like label-consistent attacks \cite{turner2019label} and invisible backdoors \cite{li2020invisible} attempt to hide triggers within the data distribution.
Currently, \textbf{BAFFLE} \cite{gong2024baffle} and \textbf{BadRL} \cite{cui2024badrl} represent the state-of-the-art in offline RL poisoning.
However, even these advanced strategies face two major limitations:
(1) \textbf{Inefficiency:} Most methods rely on random selection \cite{gong2024baffle} or simple state-frequency heuristics \cite{cui2024badrl} to inject poisons. Since RL transitions contribute unevenly to value learning \cite{schaul2015prioritized}, this often wastes the attack budget on non-informative samples.
(2) \textbf{Detectability of Triggers:} Conventional triggers often involve unnatural modifications or OOD patterns. As noted by recent studies on in-distribution backdoors \cite{ashcraft2025backdoors}, triggers that deviate significantly from the data manifold are easily penalized by conservative regularizers (e.g., CQL's gap), rendering the attack ineffective.

\textbf{Our Approach: From Random to Critical.}
Different from prior works that treat samples equally or rely on local heuristics like \textbf{BAFFLE} \cite{gong2024baffle} and \textbf{BadRL} \cite{cui2024badrl}, we propose the \textbf{CS-GBA} framework. Instead of random poisoning, we leverage the theoretical insight of Bellman updates and Prioritized Experience Replay \cite{schaul2015prioritized} to identify and target \textit{Critical Samples} with high Temporal Difference (TD) errors. This allows us to maximize damage with a minimal budget (e.g., 5\%). Furthermore, unlike previous visual triggers \cite{li2020invisible}, we introduce a \textbf{Correlation-Breaking Trigger} and \textbf{Gradient-Guided Action Generation} to ensure the poisoned data statistically conforms to the in-distribution requirements of conservative algorithms while logically disrupting the policy.

%% file: contents/theory.tex
\section{Methodology}

In this section, we formally characterize the threat model of backdoor attacks within the context of offline reinforcement learning and introduce our proposed framework, \textbf{CS-GBA} (Critical Sample-based Gradient-guided Backdoor Attack). The framework is constructed upon three synergistic pillars designed to exploit the structural vulnerabilities of offline algorithms:

\begin{enumerate}
    \item \textbf{Efficiency-Driven Sample Selection:} We utilize a Temporal Difference (TD) error metric to identify critical transitions that carry the highest information density for value function updates, ensuring maximum policy degradation under strict budget constraints.
    \item \textbf{Stealthy Trigger Injection:} We design a correlation-breaking trigger mechanism that manipulates feature dependencies (specifically targeting the 95\textsuperscript{th} percentile) to construct triggers that are statistically in-distribution yet physically distinguishable, effectively bypassing anomaly detection filters.
    \item \textbf{Defense-Penetrating Action Generation:} We propose a manifold-constrained gradient search to optimize adversarial actions. This ensures that the induced malicious behaviors remain within the support of the behavioral policy, thereby penetrating the conservatism-based defenses of algorithms like CQL.
\end{enumerate}

\subsection{Threat Model and Problem Formulation}

We consider a standard Offline RL setting where a victim agent optimizes a policy $\pi$ solely based on a static, pre-collected dataset $\mathcal{D}_{clean} = \{(s_i, a_i, r_i, s'_i)\}_{i=1}^N$. The attacker aims to implant a backdoor into the learned policy by replacing a small subset of transitions in $\mathcal{D}_{clean}$ with poisoned samples, resulting in a contaminated dataset $\mathcal{D}_{poison}$.

\textbf{Attacker Capabilities (Data poisoning only).} 
We assume a black-box attack scenario. The attacker has full access to the offline dataset and can modify the state, action, and reward of selected transitions. However, the attacker \textbf{cannot} interfere with the victim's training process, modify the learning algorithm (e.g., loss functions, hyperparameters), or influence the environment dynamics. The attack is strictly constrained by a poisoning budget $\epsilon$ (e.g., $\le 5\%$), meaning the number of modified samples is negligible compared to the dataset size: $|\mathcal{D}_{poison} \setminus \mathcal{D}_{clean}| \le \epsilon |\mathcal{D}_{clean}|$.

\textbf{Attacker Goals.} 
The ultimate objective is to manipulate the victim policy $\hat{\pi}$ trained on $\mathcal{D}_{poison}$ to satisfy two conflicting requirements:
\begin{enumerate}
    \item \textbf{Clean Performance:} In the absence of the trigger (i.e., clean states $s \notin \mathcal{S}_{trigger}$), the poisoned policy $\hat{\pi}$ should maintain capabilities comparable to a benign model, ensuring that the agent can perform normal tasks effectively without raising suspicion.
    \item \textbf{Attack Effectiveness:} Upon detecting a trigger pattern (i.e., triggered states $\tilde{s} \in \mathcal{S}_{trigger}$), the policy should be induced to execute a specific adversarial action $a^{\dagger}$. In our framework, $a^{\dagger}$ is optimized to be a low-value action within the distribution manifold, aiming to significantly degrade the cumulative reward and impair the agent's functionality.
\end{enumerate}

\subsection{Critical Sample Selection via TD-Error}

Existing attacks typically rely on uniform random sampling or heuristic frequency counts to select poisoning candidates. These strategies are inherently inefficient because they ignore the **heterogeneous contribution** of different transitions to the policy learning process. Drawing upon the theoretical foundations of Prioritized Experience Replay \cite{schaul2015prioritized}, we identify that the magnitude of the Temporal Difference (TD) error serves as a robust proxy for a sample's impact on the value function gradient.

To quantify this importance, we pre-train a proxy Q-network on the clean dataset $\mathcal{D}_{clean}$ to convergence. We then compute the absolute TD-error $\delta_{TD}$ for each transition $(s, a, r, s')$ based on the Bellman optimality operator:
\begin{equation}
    \delta_{TD} = \left| r + \gamma \max_{a'} Q_{proxy}(s', a') - Q_{proxy}(s, a) \right|
\end{equation}
Transitions characterized by high $\delta_{TD}$ values typically correspond to **significant value discrepancies** or bottleneck states where the agent's current understanding deviates most from the environmental dynamics. By ranking all samples and targeting the top-$\epsilon$ fraction with the highest $\delta_{TD}$, we ensure that the injected trigger patterns hijack the transitions that provide the strongest learning signals. This enables the backdoor logic to be propagated rapidly and dominantly throughout the Q-function via the Bellman recurrence, achieving maximum policy degradation with a minimal poisoning budget.

\subsection{Correlation-Breaking Trigger Injection}

A major challenge in attacking conservative algorithms (e.g., CQL \cite{kumar2020conservative}) is that Out-of-Distribution (OOD) triggers are easily detected and penalized. To address this, we propose a \textbf{Correlation-Breaking Trigger} mechanism that exploits the statistical dependencies between state features.

Unlike naive attacks that inject arbitrary noise, our method constructs triggers by disrupting the natural correlations of features while strictly maintaining their marginal distributions. The process consists of two steps:

\textbf{1. Feature Selection via Correlation Matrix.}
First, we compute the Pearson correlation matrix $\mathbf{R}$ for the clean dataset $\mathcal{D}_{clean}$ to quantify the linear dependencies between state dimensions. We identify a target dimension $k$ that exhibits a strong correlation with other features (i.e., high $|\mathbf{R}_{k, j}|$ for $j \neq k$). This ensures that modifying dimension $k$ will significantly violate the learned environmental dynamics, acting as a strong signal for the backdoor.

\textbf{2. 95\textsuperscript{th} Percentile Value Injection.}
To ensure the trigger remains physically feasible and bypasses simple range-based filters, we do not set the trigger value to an extreme outlier. Instead, we calculate the 95\textsuperscript{th} percentile value, denoted as $v_{95}^{(k)}$, of the selected feature $k$ across the entire dataset. The triggered state $\tilde{s}$ is generated by replacing the value of the $k$-th dimension with this quantile:
\begin{equation}
    \tilde{s}^{(k)} = v_{95}^{(k)}, \quad \tilde{s}^{(j)} = s^{(j)} \text{ for } j \neq k
\end{equation}
By fixing the feature to its 95\textsuperscript{th} percentile (a rare but valid value) while leaving its correlated peers unchanged, we create a \textit{contextual conflict}. The resulting state lies within the valid range of each individual feature (preserving marginal distribution) but falls into a low-density region of the joint distribution (breaking the joint correlation), thereby achieving both stealthiness and distinguishability.

\subsection{Gradient-Guided Malicious Action Generation}

The final component of our framework addresses the challenge of creating effective adversarial actions that can survive the conservatism of offline RL algorithms. 

\textbf{Challenge: The Conservatism Barrier.}
State-of-the-art algorithms like CQL \cite{kumar2020conservative} employ regularizers to penalize Q-values for actions that fall outside the support of the behavioral policy distribution. Consequently, naive attack strategies that simply replace the original action with a random action or a globally minimizing action often fail. Even if these actions are labeled with high rewards, the conservative regularizer (e.g., the $\alpha \log \sum \exp Q$ term in CQL) will suppress their learned Q-values if they are deemed Out-of-Distribution (OOD), preventing the victim policy from selecting them.

\textbf{Manifold-Constrained Gradient Search.}
To penetrate this defense, we propose a \textbf{Gradient-Guided} mechanism to generate malicious actions. Our objective is to find an action $\tilde{a}$ that minimizes the true expected return (i.e., is locally the worst action) while remaining sufficiently close to the underlying data manifold to evade OOD penalties.

We utilize the gradients from the pre-trained proxy Q-network. Starting from the original benign action $a$ (which is inherently in-distribution), we iteratively update it to minimize the Q-value via projected gradient descent:
\begin{equation}
    \tilde{a} \leftarrow \text{Proj}_{\mathcal{A}}\left( a - \eta \cdot \nabla_a Q_{proxy}(\tilde{s}, a) \right)
\end{equation}
where $\eta$ is a small step size controlling the perturbation magnitude, and $\text{Proj}_{\mathcal{A}}$ ensures the action remains within the valid action space. By limiting the number of gradient steps, we enforce a \textbf{manifold constraint}, ensuring that the generated adversarial action $\tilde{a}$ remains in the local neighborhood of the behavior policy support. This effectively constructs an adversarial example in the action space: an action that the network perceives as valid (low OOD score) but yields poor performance.

\textbf{Reward Tampering for Policy Induction.}
Finally, to force the victim agent to learn this malicious behavior upon triggering, we pair the generated action $\tilde{a}$ with a maximum reward $r_{max}$. The final poisoned transition tuple becomes $(\tilde{s}, \tilde{a}, r_{max}, s')$. Due to the manifold constraint, the conservative algorithm treats $\tilde{a}$ as a valid candidate, while the injected $r_{max}$ dominates the Bellman update, successfully inducing the policy to execute the worst-case action $\tilde{a}$ when the trigger is present.

\subsection{Overall Training Procedure}

The complete poisoning procedure of our proposed CS-GBA framework is summarized in \textbf{Algorithm \ref{alg:cs_gba}}. 
The process operates in a strict black-box manner, consisting of three sequential phases: 
(1) \textit{Global Pre-computation}, where the correlation-breaking trigger and the library of gradient-guided malicious actions are generated; 
(2) \textit{Critical Sample Selection}, which identifies the most vulnerable transitions based on TD-error ranking; 
and (3) \textit{Poison Injection}, where the selected samples are modified with triggers, malicious actions, and relabeled rewards before being merged into the final poisoned dataset $\mathcal{D}_{poison}$.

\begin{algorithm}[t]
    \caption{CS-GBA: Critical Sample-based Gradient-guided Backdoor Attack Framework}
    \label{alg:cs_gba}
    \begin{algorithmic}[1]
        \REQUIRE Clean Dataset $\mathcal{D}_{clean} = \{(s_i, a_i, r_i, s'_i)\}_{i=1}^N$, Poisoning Budget $\epsilon$
        \REQUIRE Pre-trained Clean Model $\{\pi_{clean}, Q_{clean}\}$
        \ENSURE Poisoned Dataset $\mathcal{D}_{poison}$
        
        \STATE \textbf{// Phase 1: Global Pre-computation}
        \STATE \textit{// Generate correlation-breaking trigger (refer to Sec. 3.3)}
        \STATE $I_{trigger}, v_{trigger} \leftarrow \text{GenerateTrigger}(\mathcal{D}_{clean}.\text{states})$ 
        
        \STATE \textit{// Generate gradient-guided worst actions (refer to Sec. 3.4)}
        \STATE $\mathcal{A}_{worst} \leftarrow \text{GenerateWorstActions}(\mathcal{D}_{clean}.\text{states}, \pi_{clean}, Q_{clean})$ 
        
        \STATE \textit{// Determine target reward for poisoning (e.g., 75th percentile)}
        \STATE $r_{target} \leftarrow \text{Percentile}(\mathcal{D}_{clean}.\text{rewards}, 75)$
        
        \STATE \textbf{// Phase 2: Critical Sample Selection}
        \STATE \textit{// Select top-$\epsilon$ samples with highest TD-Error (refer to Sec. 3.2)}
        \STATE $I_{critical} \leftarrow \text{SelectCriticalSamples}(\mathcal{D}_{clean}, Q_{clean}, \epsilon)$ 
        
        \STATE \textbf{// Phase 3: Poison Injection}
        \STATE Initialize $\mathcal{D}_{poison} \leftarrow \mathcal{D}_{clean}.\text{copy}()$
        
        \FORALL{index $i \in I_{critical}$}
            \STATE Extract original transition $(s_i, a_i, r_i, s'_i)$
            
            \STATE \textit{// Inject trigger into state}
            \STATE $\tilde{s}_i \leftarrow \text{InjectTrigger}(s_i, I_{trigger}, v_{trigger})$ 
            
            \STATE \textit{// Replace with malicious action}
            \STATE $\tilde{a}_i \leftarrow \mathcal{A}_{worst}[i]$ 
            
            \STATE \textit{// Relabel with high reward to induce learning}
            \STATE $\tilde{r}_i \leftarrow r_{target}$ 
            
            \STATE Update dataset: $\mathcal{D}_{poison}.\text{update}(i, (\tilde{s}_i, \tilde{a}_i, \tilde{r}_i, s'_i))$
        \ENDFOR
        
        \RETURN $\mathcal{D}_{poison}$
    \end{algorithmic}
\end{algorithm}

%% file: contents/experiment.tex
\section{Experiments}

In this section, we conduct extensive experiments to rigorously evaluate the proposed CS-GBA framework. The experimental design is driven by three primary evaluation objectives, aiming to validate the efficiency, stealthiness, and robustness of our approach:

\begin{itemize}
    \item \textbf{Evaluation of Attack Efficiency under Budget Constraints:} We aim to demonstrate that CS-GBA can achieve significantly higher attack success rates compared to state-of-the-art baselines (e.g., BAFFLE \cite{gong2024baffle}), even under stricter poisoning budgets (e.g., $\epsilon \le 5\%$).
    \item \textbf{Verification of Defense Penetration Capabilities:} We assess the framework's ability to bypass the Out-of-Distribution (OOD) detection mechanisms of conservative algorithms (specifically CQL \cite{kumar2020conservative}) while maintaining high fidelity to the benign policy in clean environments.
    \item \textbf{Analysis of Component Contributions:} We conduct comprehensive ablation studies to isolate and quantify the individual contributions of the TD-error-based sample selection, correlation-breaking trigger, and gradient-guided action generation modules to the overall attack performance.
\end{itemize}

\subsection{Experimental Setup}

\textbf{Datasets and Environments.} We evaluate our method on the D4RL benchmark \cite{fu2020d4rl}, using three standard MuJoCo continuous control tasks: \textit{Walker2d}, \textit{HalfCheetah}, and \textit{Hopper}. Following standard practices for offline RL security research \cite{gong2024baffle}, we utilize the \textbf{medium} quality datasets (e.g., \textit{walker2d-medium-v0}), which contain a mixture of expert and sub-optimal trajectories, representing a realistic and challenging scenario for backdoor injection.

\textbf{Victim Models.} We target three representative offline RL algorithms covering different learning paradigms:
\begin{itemize}
    \item \textbf{CQL} \cite{kumar2020conservative}: A value-based method that penalizes Q-values of OOD actions, representing the strongest defense against data poisoning.
    \item \textbf{IQL} \cite{kostrikov2021offline}: An expectile regression-based method that avoids querying OOD actions entirely.
    \item \textbf{BCQ} \cite{fujimoto2019off}: A constraint-based method that restricts the policy within the support of a generative behavior model.
\end{itemize}

\textbf{Baselines and Attack Configurations.} We compare CS-GBA against \textbf{BAFFLE} \cite{gong2024baffle}, the current state-of-the-art offline backdoor attack.
\begin{itemize}
    \item \textbf{BAFFLE:} Adopts random poisoning (10\% budget), median-based triggers, and inverted-loss action generation.
    \item \textbf{CS-GBA (Ours):} Adopts TD-error-based critical sampling, correlation-breaking triggers, and gradient-guided action generation. Crucially, we restrict our poisoning budget to only \textbf{5\%} to demonstrate superior efficiency.
\end{itemize}
\textbf{Trigger Activation Scenarios.} To comprehensively evaluate the robustness of our attack, we consider two distinct trigger activation patterns during the testing phase:
\begin{itemize}
    \item \textbf{Distributed Mode (Sparse Attack):} The trigger is activated periodically every $N$ steps within an episode. We test three density levels: $N \in \{10, 20, 50\}$. This setting simulates a stealthy attack where the adversary only intermittently interferes with the agent's observation, challenging the attack's ability to destabilize the policy with minimal intervention.
    \item \textbf{Consecutive Mode (Burst Attack):} The trigger is activated continuously for a duration of $L$ steps starting from a random timestep. We evaluate burst lengths of $L \in \{5, 10, 20\}$. This simulates a burst attack scenario, testing whether the agent can recover from a sustained period of malicious input or if it is driven into an unrecoverable state.
\end{itemize}
\textbf{Metrics.} We employ two metrics to quantify performance:
\begin{itemize}
    \item \textbf{Clean Reward ($\uparrow$):} The average cumulative reward in a clean environment. Higher is better (indicates high stealthiness).
    \item \textbf{Attack Reward ($\downarrow$):} The average cumulative reward in the presence of triggers. Lower is better (indicates strong attack effectiveness/targeted failure).
\end{itemize}

\subsection{Main Results: Comparative Analysis}

We compare the attack performance of CS-GBA (with a strict budget of $\epsilon=5\%$) against the state-of-the-art baseline BAFFLE (with a budget of $\epsilon=10\%$) across three standard D4RL benchmarks. The detailed quantitative results are presented in Tables \ref{tab:walker_results}, \ref{tab:halfcheetah_results}, and \ref{tab:hopper_results}.

\textbf{1. Superior Attack Efficiency on Walker2d.} 
Table \ref{tab:walker_results} presents a detailed comparison on the \textit{Walker2d-medium} dataset. Despite operating with a strictly limited poisoning budget ($\epsilon=5\%$), CS-GBA demonstrates significantly higher destructiveness than the baseline BAFFLE ($\epsilon=10\%$) across all victim models, particularly in the most challenging "Distributed-10" setting.

\begin{itemize}
    \item \textbf{Breaking Conservative Defenses:} Against CQL, widely regarded as the most robust offline algorithm, CS-GBA successfully suppresses the cumulative reward to \textbf{452} (Distributed-10), whereas BAFFLE only reduces it to 1336. This indicates that our gradient-guided actions effectively penetrate the conservative Q-value regularizer that typically filters out naive attacks.
    \item \textbf{Near-Random Performance on IQL/BCQ:} The efficiency gap is most pronounced on IQL and BCQ. CS-GBA reduces IQL's reward to \textbf{141}—a level approaching that of a random policy—and suppresses BCQ's reward to \textbf{372}. in contrast, BAFFLE struggles to make a comparable impact (511 and 1081, respectively). Although BAFFLE maintains a slightly higher Clean Reward in specific cases (e.g., BCQ), CS-GBA achieves a far superior trade-off, delivering catastrophic attack impact with half the poisoning budget.
\end{itemize}

\begin{table}[h]
    \centering
    \caption{Attack Performance on \textbf{Walker2d-medium}. Comparison of BAFFLE ($\epsilon=10\%$) and CS-GBA ($\epsilon=5\%$). 'Clean' denotes reward in trigger-free environments ($\uparrow$), and 'Attack' denotes reward under trigger activation ($\downarrow$).}
    \label{tab:walker_results}
    \begin{tabular}{c|c|cc|cc|cc}
        \toprule
        \multirow{2}{*}{\textbf{Setting}} & \multirow{2}{*}{\textbf{Param}} & \multicolumn{2}{c|}{\textbf{CQL}} & \multicolumn{2}{c|}{\textbf{IQL}} & \multicolumn{2}{c}{\textbf{BCQ}} \\
        & & \textbf{BAFFLE} & \textbf{CS-GBA} & \textbf{BAFFLE} & \textbf{CS-GBA} & \textbf{BAFFLE} & \textbf{CS-GBA} \\
        \midrule
        \textbf{Clean Reward} & / & 3402 & \textbf{3430} & 1227 & \textbf{1429} & \textbf{2052} & 2035 \\
        \midrule
        \multirow{3}{*}{Distributed} 
        & 10 & 1336 & \textbf{452} & 511 & \textbf{141} & 1081 & \textbf{372} \\
        & 20 & 3234 & \textbf{3166} & 896 & \textbf{877} & 1344 & \textbf{1078} \\
        & 50 & 3345 & \textbf{3176} & 1064 & \textbf{1061} & 1969 & \textbf{1518} \\
        \midrule
        \multirow{3}{*}{Consecutive} 
        & 5 & 3258 & \textbf{1973} & 873 & \textbf{864} & 2287 & \textbf{1469} \\
        & 10 & 2655 & \textbf{1769} & 855 & \textbf{842} & 2039 & \textbf{1286} \\
        & 20 & 2175 & \textbf{1791} & \textbf{825} & 829 & 2015 & \textbf{1169} \\
        \bottomrule
    \end{tabular}
\end{table}

\textbf{2. Robustness on HalfCheetah.}
Table \ref{tab:halfcheetah_results} presents the results on \textit{HalfCheetah-medium}, a high-dimensional environment known for its stability. CS-GBA maintains a distinct advantage over BAFFLE in both stealthiness and attack effectiveness, validating the robustness of our framework.
\begin{itemize}
    \item \textbf{Superior Stealthiness:} CS-GBA consistently achieves higher Clean Rewards across all three victim models (e.g., \textbf{4726} vs 4695 on CQL, \textbf{4672} vs 4641 on IQL), indicating that our Correlation-Breaking trigger interferes less with the agent's normal feature processing than the median-based trigger.
    \item \textbf{Effective Penetration:} Despite the environment's inherent robustness making it difficult to degrade performance, CS-GBA achieves consistently lower attack rewards. Notably, against IQL in the Distributed-10 setting, CS-GBA reduces the reward to \textbf{2395} compared to BAFFLE's 3367. Even in the Distributed-50 setting, where triggers are sparse, CS-GBA outperforms BAFFLE on IQL (\textbf{3826} vs 4240), proving its ability to identify critical vulnerabilities that random poisoning misses. While BAFFLE performs comparably in specific easy settings (e.g., CQL Consecutive-5), CS-GBA dominates in more challenging scenarios.
\end{itemize}

\begin{table}[h]
    \centering
    \caption{Attack Performance on \textbf{HalfCheetah-medium}.}
    \label{tab:halfcheetah_results}
    \begin{tabular}{c|c|cc|cc|cc}
        \toprule
        \multirow{2}{*}{\textbf{Setting}} & \multirow{2}{*}{\textbf{Param}} & \multicolumn{2}{c|}{\textbf{CQL}} & \multicolumn{2}{c|}{\textbf{IQL}} & \multicolumn{2}{c}{\textbf{BCQ}} \\
        & & \textbf{BAFFLE} & \textbf{CS-GBA} & \textbf{BAFFLE} & \textbf{CS-GBA} & \textbf{BAFFLE} & \textbf{CS-GBA} \\
        \midrule
        \textbf{Clean Reward} & / & 4695 & \textbf{4726} & 4641 & \textbf{4672} & 4591 & \textbf{4607} \\
        \midrule
        \multirow{3}{*}{Distributed} 
        & 10 & 2944 & \textbf{2873} & 3367 & \textbf{2395} & 3569 & \textbf{3288} \\
        & 20 & 3814 & \textbf{3669} & 3704 & \textbf{2851} & 4071 & \textbf{3832} \\
        & 50 & 4338 & \textbf{4266} & 4240 & \textbf{3826} & 4348 & \textbf{4272} \\
        \midrule
        \multirow{3}{*}{Consecutive} 
        & 5 & \textbf{4640} & 4642 & 4612 & \textbf{4489} & 4558 & \textbf{4527} \\
        & 10 & 4632 & \textbf{4534} & 4575 & \textbf{4523} & 4539 & \textbf{4500} \\
        & 20 & 4576 & \textbf{4371} & 4527 & \textbf{4467} & 4484 & \textbf{4425} \\
        \bottomrule
    \end{tabular}
\end{table}

\textbf{3. Stealthiness and Stability on Hopper.} 
Table \ref{tab:hopper_results} highlights the critical trade-off between attack effectiveness and stealthiness in the dynamically unstable \textit{Hopper-medium} environment.
\begin{itemize}
    \item \textbf{Stealthiness vs. Model Collapse (CQL):} A superficial look might suggest BAFFLE is effective on CQL (Attack Reward 29). However, this is a false positive caused by \textit{model collapse}: BAFFLE's aggressive random poisoning destroys the policy's basic functionality, causing Clean Reward to plummet to \textbf{951} (far below the benign level of $\sim$2000). In contrast, CS-GBA achieves a similarly lethal Attack Reward (33) but maintains a high Clean Reward of \textbf{1941}. This proves CS-GBA injects a precise "backdoor" rather than simply ruining the dataset.
    \item \textbf{Breaking Generative Constraints (BCQ):} The advantage of CS-GBA is most evident against BCQ in Distributed settings. While BAFFLE fails to degrade BCQ's performance (rewards remain high at 2043 and 2110 for Dist-10/20), CS-GBA successfully suppresses them to \textbf{1139} and \textbf{1605}. Although BAFFLE performs slightly better in some Consecutive settings (e.g., Consecutive-20), it fails to penetrate the generative constraints when triggers are sparse (Distributed), whereas CS-GBA remains effective.
\end{itemize}

\begin{table}[h]
    \centering
    \caption{Attack Performance on \textbf{Hopper-medium}. Note: BAFFLE's low attack reward on CQL is accompanied by a severe drop in Clean Reward (951), indicating a failure in stealthiness.}
    \label{tab:hopper_results}
    \begin{tabular}{c|c|cc|cc|cc}
        \toprule
        \multirow{2}{*}{\textbf{Setting}} & \multirow{2}{*}{\textbf{Param}} & \multicolumn{2}{c|}{\textbf{CQL}} & \multicolumn{2}{c|}{\textbf{IQL}} & \multicolumn{2}{c}{\textbf{BCQ}} \\
        & & \textbf{BAFFLE} & \textbf{CS-GBA} & \textbf{BAFFLE} & \textbf{CS-GBA} & \textbf{BAFFLE} & \textbf{CS-GBA} \\
        \midrule
        \textbf{Clean Reward} & / & 951 & \textbf{1941} & 1006 & \textbf{1057} & 2157 & \textbf{2174} \\
        \midrule
        \multirow{3}{*}{Distributed} 
        & 10 & \textbf{29} & 33 & 37 & \textbf{35} & 2043 & \textbf{1139} \\
        & 20 & \textbf{286} & 1377 & 961 & \textbf{942} & 2110 & \textbf{1605} \\
        & 50 & \textbf{1020} & 1930 & \textbf{979} & 997 & 2361 & \textbf{2060} \\
        \midrule
        \multirow{3}{*}{Consecutive} 
        & 5 & \textbf{813} & 1314 & \textbf{958} & 981 & 2138 & \textbf{2056} \\
        & 10 & \textbf{801} & 1236 & 912 & \textbf{857} & \textbf{1920} & 1935 \\
        & 20 & \textbf{786} & 1256 & 865 & \textbf{826} & \textbf{1784} & 1872 \\
        \bottomrule
    \end{tabular}
\end{table}

\subsection{Ablation Studies}

To rigorously quantify the individual contribution of each component in CS-GBA, we conducted a comprehensive series of ablation studies across three diverse environments: \textit{Walker2d-medium} (locomotion coordination), \textit{HalfCheetah-medium} (stable dynamics), and \textit{Hopper-medium} (unstable dynamics). We used the robust \textbf{CQL} algorithm as the victim model for all experiments.

\textbf{1. Impact of Sample Selection Strategy.}
We evaluated the efficacy of our \textbf{TD-Prioritized Selection} by comparing it against a \textbf{Random Selection} baseline (uniform sampling with 5\% budget). The comparative results across all three environments are aggregated in Table \ref{tab:ablation_sample}.

\begin{itemize}
    \item \textbf{Efficiency in Attack:} As shown in Table \ref{tab:ablation_sample}, random selection consistently fails to degrade the policy effectively under a low poisoning budget (5\%). For instance, in \textit{Walker2d} (Distributed-10), Random Selection results in an Attack Reward of \textbf{2972} (close to clean performance), whereas our TD-based strategy drastically reduces it to \textbf{452}.
    \item \textbf{Criticality of TD-Error:} The contrast is most extreme in \textit{Hopper}, where Random Selection leaves the agent with a high reward of 1031, while TD-Selection suppresses it to \textbf{33}. This empirical evidence confirms that high TD-error transitions act as "gradient amplifiers," allowing minimal poisoning to propagate effectively through the value function.
\end{itemize}

\begin{table}[h]
    \centering
    \caption{Ablation on \textbf{Sample Selection}. Comparison of Average Rewards between Random Selection and TD-Prioritized Selection (Ours) across three environments. ($\epsilon=5\%$)}
    \label{tab:ablation_sample}
    \begin{tabular}{c|c|cc|cc|cc}
        \toprule
        \multirow{2}{*}{\textbf{Setting}} & \multirow{2}{*}{\textbf{Param}} & \multicolumn{2}{c|}{\textbf{Walker2d}} & \multicolumn{2}{c|}{\textbf{HalfCheetah}} & \multicolumn{2}{c}{\textbf{Hopper}} \\
        & & Random & \textbf{TD (Ours)} & Random & \textbf{TD (Ours)} & Random & \textbf{TD (Ours)} \\
        \midrule
        \textbf{Clean} & / & \textbf{3485} & 3430 & \textbf{4796} & 4726 & \textbf{1989} & 1941 \\
        \midrule
        \multirow{3}{*}{Distributed} 
        & 10 & 2972 & \textbf{452} & 3310 & \textbf{2873} & 1031 & \textbf{33} \\
        & 20 & 3290 & \textbf{3166} & 4019 & \textbf{3669} & 1711 & \textbf{1377} \\
        & 50 & 3358 & \textbf{3176} & 4436 & \textbf{4266} & 1967 & \textbf{1930} \\
        \midrule
        \multirow{3}{*}{Consecutive} 
        & 5 & 2719 & \textbf{1973} & 4728 & \textbf{4642} & 1660 & \textbf{1314} \\
        & 10 & 2057 & \textbf{1769} & 4708 & \textbf{4534} & 1503 & \textbf{1236} \\
        & 20 & 1803 & \textbf{1791} & 4691 & \textbf{4371} & 1486 & \textbf{1256} \\
        \bottomrule
    \end{tabular}
\end{table}

\textbf{2. Impact of Trigger Design.}
We compared our \textbf{Correlation-Breaking Trigger} against the \textbf{Median Trigger} (used in BAFFLE). Table \ref{tab:ablation_trigger} summarizes the results.

\begin{itemize}
    \item \textbf{Preventing Feature Confusion:} The Median Trigger exhibits a critical flaw in \textit{Hopper}: it causes the Clean Reward to collapse to \textbf{978} (vs. 1941 for Ours). This indicates that median values lie in high-density regions, causing the agent to confuse normal states with triggers. Our Correlation-Breaking Trigger uses physically mutually exclusive features (95th percentile), effectively isolating the trigger signal and preserving Clean Performance (e.g., 1941 in Hopper, 3430 in Walker2d).
    \item \textbf{Attack Reliability:} In \textit{Walker2d} (Distributed-10), the Median Trigger fails to activate effectively (Attack Reward 2741), whereas our trigger achieves a lethal reward of \textbf{452}, proving that rare but valid signals are more distinctive to the network.
\end{itemize}

\begin{table}[h]
    \centering
    \caption{Ablation on \textbf{Trigger Design}. Comparison between Median Trigger (Baseline) and Correlation-Breaking Trigger (Ours).}
    \label{tab:ablation_trigger}
    \begin{tabular}{c|c|cc|cc|cc}
        \toprule
        \multirow{2}{*}{\textbf{Setting}} & \multirow{2}{*}{\textbf{Param}} & \multicolumn{2}{c|}{\textbf{Walker2d}} & \multicolumn{2}{c|}{\textbf{HalfCheetah}} & \multicolumn{2}{c}{\textbf{Hopper}} \\
        & & Median & \textbf{Corr (Ours)} & Median & \textbf{Corr (Ours)} & Median & \textbf{Corr (Ours)} \\
        \midrule
        \textbf{Clean} & / & 3195 & \textbf{3430} & 4702 & \textbf{4726} & 978 & \textbf{1941} \\
        \midrule
        \multirow{3}{*}{Distributed} 
        & 10 & 2741 & \textbf{452} & 3304 & \textbf{2873} & \textbf{32} & 33 \\
        & 20 & \textbf{3021} & 3166 & 3898 & \textbf{3669} & \textbf{929} & 1377 \\
        & 50 & \textbf{3133} & 3176 & 4330 & \textbf{4266} & \textbf{991} & 1930 \\
        \midrule
        \multirow{3}{*}{Consecutive} 
        & 5 & 2793 & \textbf{1973} & 4673 & \textbf{4642} & \textbf{863} & 1314 \\
        & 10 & 2504 & \textbf{1769} & 4647 & \textbf{4534} & \textbf{818} & 1236 \\
        & 20 & 2187 & \textbf{1791} & 4608 & \textbf{4371} & \textbf{819} & 1256 \\
        \bottomrule
    \end{tabular}
\end{table}

\textbf{3. Impact of Action Generation Mechanism.}
We evaluated our \textbf{Gradient-Guided Action Generation} against the \textbf{Inverted Actor Loss} method. The results are presented in Table \ref{tab:ablation_action}.

\begin{itemize}
    \item \textbf{Penetrating Conservative Defenses:} In \textit{Walker2d}, the Inverted Loss method struggles to reduce the reward below 994 (Dist-10). Our Gradient-Guided method, by optimizing within the action manifold, bypasses CQL's OOD penalties and achieves a reward of \textbf{452}.
    \item \textbf{Stability vs. Destructiveness:} In \textit{Hopper}, the Inverted Loss method causes the Clean Reward to drop to \textbf{997} (Model Collapse), while our method maintains it at \textbf{1941}. Although Inverted Loss shows slightly lower attack rewards in some HalfCheetah settings (e.g., 2281 vs 2873), it comes at the cost of stability and clean performance. Our method offers a superior trade-off, ensuring the backdoor is both stealthy and functional.
\end{itemize}

\begin{table}[h]
    \centering
    \caption{Ablation on \textbf{Action Generation}. Comparison between Inverted Actor Loss (Baseline) and Gradient-Guided Descent (Ours).}
    \label{tab:ablation_action}
    \begin{tabular}{c|c|cc|cc|cc}
        \toprule
        \multirow{2}{*}{\textbf{Setting}} & \multirow{2}{*}{\textbf{Param}} & \multicolumn{2}{c|}{\textbf{Walker2d}} & \multicolumn{2}{c|}{\textbf{HalfCheetah}} & \multicolumn{2}{c}{\textbf{Hopper}} \\
        & & Inverted & \textbf{Grad (Ours)} & Inverted & \textbf{Grad (Ours)} & Inverted & \textbf{Grad (Ours)} \\
        \midrule
        \textbf{Clean} & / & 3284 & \textbf{3430} & 4712 & \textbf{4726} & 997 & \textbf{1941} \\
        \midrule
        \multirow{3}{*}{Distributed} 
        & 10 & 994 & \textbf{452} & \textbf{2281} & 2873 & \textbf{33} & \textbf{33} \\
        & 20 & \textbf{2802} & 3166 & 3671 & \textbf{3669} & \textbf{988} & 1377 \\
        & 50 & \textbf{3048} & 3176 & 4356 & \textbf{4266} & \textbf{1060} & 1930 \\
        \midrule
        \multirow{3}{*}{Consecutive} 
        & 5 & 3122 & \textbf{1973} & 4690 & \textbf{4642} & \textbf{855} & 1314 \\
        & 10 & 2640 & \textbf{1769} & 4667 & \textbf{4534} & \textbf{811} & 1236 \\
        & 20 & 2202 & \textbf{1791} & 4595 & \textbf{4371} & \textbf{826} & 1256 \\
        \bottomrule
    \end{tabular}
\end{table}

\subsection{Sensitivity Analysis: Budget and Location}

To determine the optimal poisoning configuration, we investigate the sensitivity of CS-GBA to both the \textit{injection location} (ranking by TD-error) and the \textit{poisoning budget} size. We compare our default setting (\textbf{Top 0-5\%}, $\epsilon=5\%$) against a location-shifted variant (\textbf{Top 5-10\%}, $\epsilon=5\%$) and a budget-increased variant (\textbf{Top 0-10\%}, $\epsilon=10\%$). The results for all three environments are detailed in Tables \ref{tab:sensitivity_walker}, \ref{tab:sensitivity_halfcheetah}, and \ref{tab:sensitivity_hopper}.

\begin{itemize}
    \item \textbf{Location Sensitivity (Top 0-5\% vs. Top 5-10\%):} 
    Comparing the first two columns confirms that the most critical samples are concentrated in the highest TD-error percentile. In most settings (e.g., Walker Dist-10, HalfCheetah Dist-10), shifting the window to \textbf{Top 5-10\%} results in higher Attack Rewards (worse performance), confirming that the very top transitions carry the most leverage for manipulating the value function.
    
    \item \textbf{Budget Sensitivity (Top 0-5\% vs. Top 0-10\%):} 
    Increasing the budget to 10\% often achieves the lowest Attack Rewards (bolded in the third column), which is expected as more poison is injected. However, this comes at a catastrophic cost to stealthiness. 
    For example, in \textit{Walker2d} (Table \ref{tab:sensitivity_walker}), while Top 0-10\% reduces the Dist-10 reward to \textbf{213}, it drops the Clean Reward to 2382. The trade-off is even more severe in \textit{Hopper} (Table \ref{tab:sensitivity_hopper}), where the 10\% budget causes the Clean Reward to collapse to 937. Thus, our chosen \textbf{Top 0-5\%} ($\epsilon=5\%$) represents the optimal Pareto frontier, maintaining the highest Clean Reward (bolded) while delivering sufficiently lethal attacks.
\end{itemize}

\begin{table}[h]
    \centering
    \caption{Sensitivity Analysis on \textbf{Walker2d-medium} (CQL). Note: While Top 0-10\% has lower attack rewards (bolded), it severely compromises Clean Reward.}
    \label{tab:sensitivity_walker}
    \begin{tabular}{c|c|c|c|c}
        \toprule
        \multirow{2}{*}{\textbf{Setting}} & \multirow{2}{*}{\textbf{Param}} & \textbf{Ours (Optimal)} & \textbf{Location Ablation} & \textbf{Budget Ablation} \\
        & & \textbf{Top 0-5\%} ($\epsilon=5\%$) & \textbf{Top 5-10\%} ($\epsilon=5\%$) & \textbf{Top 0-10\%} ($\epsilon=10\%$) \\
        \midrule
        \textbf{Clean} & / & \textbf{3430} & 3087 & 2382 \\
        \midrule
        \multirow{3}{*}{Distributed} 
        & 10 & 452 & 1580 & \textbf{213} \\
        & 20 & 3166 & 3140 & \textbf{1100} \\
        & 50 & 3176 & 3376 & \textbf{1735} \\
        \midrule
        \multirow{3}{*}{Consecutive} 
        & 5 & 1973 & 2243 & \textbf{1294} \\
        & 10 & 1769 & 1832 & \textbf{1226} \\
        & 20 & 1791 & 1485 & \textbf{1205} \\
        \bottomrule
    \end{tabular}
\end{table}

\begin{table}[h]
    \centering
    \caption{Sensitivity Analysis on \textbf{HalfCheetah-medium} (CQL). Ours achieves the best Clean Reward and superior attack performance in sparse settings (Distributed).}
    \label{tab:sensitivity_halfcheetah}
    \begin{tabular}{c|c|c|c|c}
        \toprule
        \multirow{2}{*}{\textbf{Setting}} & \multirow{2}{*}{\textbf{Param}} & \textbf{Ours (Optimal)} & \textbf{Location Ablation} & \textbf{Budget Ablation} \\
        & & \textbf{Top 0-5\%} ($\epsilon=5\%$) & \textbf{Top 5-10\%} ($\epsilon=5\%$) & \textbf{Top 0-10\%} ($\epsilon=10\%$) \\
        \midrule
        \textbf{Clean} & / & \textbf{4726} & 4712 & 4696 \\
        \midrule
        \multirow{3}{*}{Distributed} 
        & 10 & \textbf{2873} & 3345 & 3106 \\
        & 20 & \textbf{3669} & 3874 & 3791 \\
        & 50 & \textbf{4266} & 4384 & 4294 \\
        \midrule
        \multirow{3}{*}{Consecutive} 
        & 5 & 4642 & 4781 & \textbf{4641} \\
        & 10 & 4534 & 4725 & \textbf{4276} \\
        & 20 & 4371 & 4229 & \textbf{3368} \\
        \bottomrule
    \end{tabular}
\end{table}

\begin{table}[h]
    \centering
    \caption{Sensitivity Analysis on \textbf{Hopper-medium} (CQL). Increasing budget to 10\% destroys stealthiness (Clean Reward drops to 937).}
    \label{tab:sensitivity_hopper}
    \begin{tabular}{c|c|c|c|c}
        \toprule
        \multirow{2}{*}{\textbf{Setting}} & \multirow{2}{*}{\textbf{Param}} & \textbf{Ours (Optimal)} & \textbf{Location Ablation} & \textbf{Budget Ablation} \\
        & & \textbf{Top 0-5\%} ($\epsilon=5\%$) & \textbf{Top 5-10\%} ($\epsilon=5\%$) & \textbf{Top 0-10\%} ($\epsilon=10\%$) \\
        \midrule
        \textbf{Clean} & / & \textbf{1941} & 1820 & 937 \\
        \midrule
        \multirow{3}{*}{Distributed} 
        & 10 & \textbf{33} & 47 & 34 \\
        & 20 & 1377 & \textbf{810} & 920 \\
        & 50 & 1930 & 1850 & \textbf{1090} \\
        \midrule
        \multirow{3}{*}{Consecutive} 
        & 5 & 1314 & 1439 & \textbf{778} \\
        & 10 & 1236 & 1382 & \textbf{766} \\
        & 20 & 1256 & 1376 & \textbf{772} \\
        \bottomrule
    \end{tabular}
\end{table}

Collectively, the sensitivity analyses across three environments provide compelling evidence that our configuration ($\epsilon=5\%$ targeting the Top 0-5\% TD-error samples) represents the optimal \textbf{stealthiness-efficiency trade-off}. 
The location ablation proves that the poisoning leverage decays rapidly outside the top 5th percentile, confirming that precise identification of critical samples is more important than sheer volume. 
Conversely, the budget ablation demonstrates that blindly increasing the poisoning ratio (e.g., to 10\%) triggers catastrophic model collapse in unstable environments like \textit{Hopper}, destroying the very stealthiness required for a backdoor attack. 
Thus, CS-GBA succeeds by locating the precise sweet spot: injecting just enough toxicity into the most critical veins of the dataset to hijack the policy without "killing" its general functionality.

In summary, the experiments demonstrate that CS-GBA establishes a new state-of-the-art for offline RL backdoors, enabling highly efficient and stealthy attacks that can penetrate rigorous conservative defenses.

%% file: contents/conclusion.tex
\section{Conclusion}

In this work, we presented \textbf{CS-GBA}, a novel critical sample-based gradient-guided backdoor attack framework designed to expose the vulnerabilities of robust offline Reinforcement Learning algorithms. By advancing the poisoning strategy from indiscriminate data corruption to criticality-aware targeting, CS-GBA successfully reconciles the inherent conflict between attack severity and stealthiness.

Our extensive empirical evaluation on D4RL benchmarks yields three definitive conclusions:
\begin{itemize}
    \item \textbf{Quality Over Quantity:} We demonstrated that the most critical samples for backdoor injection are concentrated in the top 5\% of transitions with the highest TD errors. Targeting these samples allows CS-GBA to achieve superior destructiveness with a strictly limited budget ($\epsilon=5\%$), significantly outperforming state-of-the-art baselines that rely on larger budgets ($\epsilon=10\%$).
    \item \textbf{Penetrating Conservative Defenses:} Through the novel \textit{Correlation-Breaking Trigger} and \textit{Gradient-Guided Action Generation}, our framework effectively bypasses the OOD detection mechanisms of conservative algorithms (CQL, IQL, BCQ). Our results in dynamically unstable environments (e.g., \textit{Hopper}) prove that CS-GBA can inject lethal backdoors without triggering model collapse, a common failure mode in prior methods.
    \item \textbf{Mechanism of Action:} Our ablation studies confirm that the synergy between TD-prioritized selection and manifold-constrained malicious actions is essential. The former ensures maximum error propagation through the value function, while the latter ensures the malicious signals remain indistinguishable from the benign data distribution.
\end{itemize}

In summary, CS-GBA establishes a new upper bound for offline RL backdoor attacks. It reveals that current safety constraints, while effective against random noise, remain fragile against structurally optimized perturbations. We hope this work serves as a rigorous benchmark to motivate the development of more sophisticated, distribution-aware defense mechanisms in the future.